\documentclass{article}
\usepackage{amssymb}

\usepackage[preprint]{corl_2025} 
\usepackage{graphicx}
\usepackage{xcolor}
\newcommand{\hakan}[1]{\textcolor{black}{#1}}
\newcommand{\hakann}[1]{\textcolor{black}{#1}}

\newcommand{\erhan}[1]{\textcolor{black}{#1}}
\usepackage{wrapfig}
\usepackage{multirow}
\usepackage{listings}
\usepackage{makecell}
\usepackage{amsmath}

\title{Neuro-Symbolic Skill Discovery for Conditional Multi-Level Planning}

\author{
  Hakan Aktas\\
  Department of Computer Science and Technology\\
  The University of Cambridge \\
  United Kingdom\\
  \texttt{hea39@cam.ac.uk} \\
  \And
  Yigit Yildirim \\
  Department of Computer Engineering\\
  Bogazici University \\
  Turkey \\
  yigit.yildirim@boun.edu.tr \\
  \And
  Ahmet Firat Gamsiz\\
  Department of Computer Engineering\\
  Bogazici University \\
  Turkey \\
  \And
  Deniz Bilge Akkoc\\
  Department of Computer Engineering\\
  Bogazici University \\
  Turkey \\
  \And
  Erhan Oztop\\
  SISREC\\
  Osaka University\\
  Japan\\
  erhan.oztop@otri.osaka-u.ac.jp\\
  \And
  Emre Ugur \\
  Department of Computer Engineering \\
  Bogazici University\\
  Turkey\\
  \texttt{emre.ugur@bogazici.edu.tr} \\
}

\begin{document}
\maketitle


\begin{abstract}
This paper proposes a novel learning architecture for acquiring generalizable high-level symbolic skills from a few unlabeled low-level skill trajectory demonstrations. The architecture involves neural networks for symbol discovery and low-level controller acquisition and a multi-level planning pipeline that utilizes the discovered symbols and the learned low-level controllers. The discovered action symbols are automatically interpreted using visual language models that are also responsible for generating high-level plans. While extracting high-level symbols, our model preserves the low-level information so that low-level action planning can be carried out by using gradient-based planning.  To assess the efficacy of our method, we tested the high and low-level planning performance of our architecture by using simulated and real-world experiments across various tasks. The experiments have shown that our method is able to manipulate objects in unseen locations and plan and execute long-horizon tasks by using novel action sequences, even in highly cluttered environments when cued by only a few demonstrations that cover small regions of the environment.

\end{abstract}

\keywords{Neuro-Symbolic Robotics, Multi-Level Planning, Skill Discovery} 


\section{Introduction} \label{section:intro}

\begin{wrapfigure}{r}{0.5\textwidth}
    \centering\includegraphics[width=0.9\linewidth]{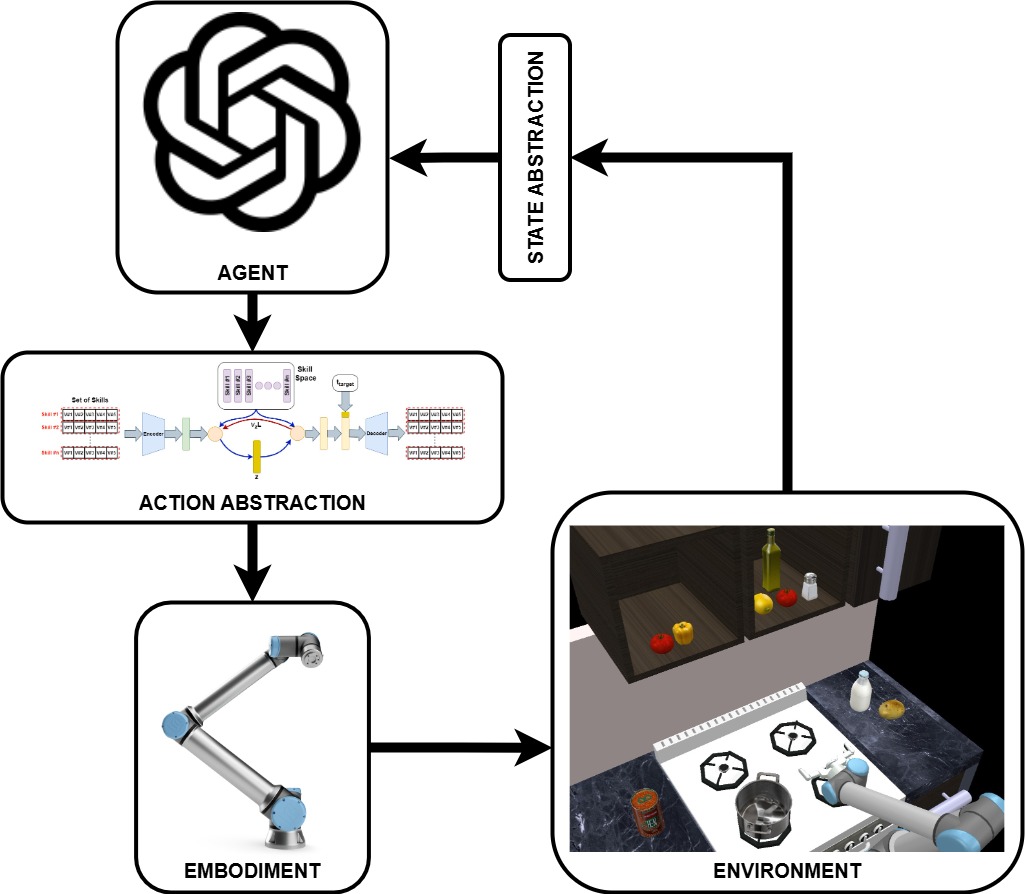}
\caption{An overview of a generalist agent that uses abstractions to interact with its environment.}
\label{fig:generalist-agent}
\end{wrapfigure}

Learning to make long-horizon plans in continuous state-action spaces, as in many real-world scenarios, is a hard problem. To address this issue, state \cite{ahmetoglu2024discovering,ahmetoglu2022deepsym,silver2023predicate}, and action abstractions \cite{konidaris2018skills} have been proposed for separating high-level decision-making from low-level perception and control. Such abstractions showed great success in making fast and effective robotic plans that can be generalized to unseen environments. 
Our paper, on the other hand, focuses on using learned action and state predicates as an intermediary layer between virtual generalist agents and environments. Learning from low-level continuous robotic data is difficult as it is complex and noisy, especially in the real world. Furthermore, the data is limited as it is not mostly produced by people and is only collected in controlled lab settings. These two issues make it hard to include raw robotic data in training datasets of multi-purpose large models such as LLMs. However, since their training data (text) provides information about numerous domains, LLMs have (although limited) reasoning capabilities in multiple domains, including reasoning for high-level robotic tasks. We believe abstractions can be used to bridge this gap between real-world environments and generalist virtual agents. State abstraction methods can be the eyes, and action abstractions can be the arms and legs of these embodied virtual generalist agents. In the case of LLMs, instead of including the robotic data in the training data of the LLM, one can train small abstraction models that could bridge the gap between the agent and its embodiment and the environment in which the embodiment resides.
This way, these abstraction models can handle all the low-level information while the agent is capable of high-level reasoning, which can be learned from text more efficiently. An overview of how a system like this would function is provided in Figure \ref{fig:generalist-agent}.

In this paper, we propose a novel action abstraction model that can learn discrete high-level skill representations from demonstrations that include low-level variations of skills. The key features of our approach are the learning of high-level skills by keeping track of the corresponding low-level information and the clustering of demonstrated skills in an unsupervised manner. We also propose a bi-level planning pipeline that an LLM agent can use to execute high-level plans. After discovering the high-level skills, the model can also be utilized to make low-level plans using a gradient-based approach and can generalize to manipulate large chunks of the environment using only several demonstrations per skill. Assuming an unlabeled demonstration dataset at the start, the proposed pipeline includes the following steps: (1)High-level skills are discovered from unlabeled demonstrations using our conditional trajectory generation model. (2) Discovered high-level skills are labeled either by an expert or a Multi-Modal LLM. (3) A new model is trained using self-supervision provided by the discovered high-level skill labels. (4) An external agent is used to make high-level plans using the discovered skills for the given goal tasks. (5) Low-level execution plans are made for each step of the high-level plan using our proposed gradient-based planning approach.

As our contribution, we propose a novel skill discovery method that can learn high-level skill representations from unlabeled trajectory demonstrations, and a bi-level planning pipeline capable of handling multi-task long-horizon plans by using the discovered symbols and skills.
\section{Related Work}\label{section:related}

Long-horizon planning in a continuous robotic environment is a challenging task. \cite{garrett2021integrated} calls this complex problem of integrating different discrete and continuous elements Task and Motion Planning (TAMP). Several studies have addressed this problem by bi-level planning, which separates tasks into two levels: ``what-to-do'' (high-level task planning) and ``how-to-do'' (low-level control). For the low-level control, \cite{silver2022learning} learns operator-specific, sub-goal-conditioned policies, which are trained using data segmented according to `switch points’ and symbolic predicate changes (assuming predicates are predefined) from preprocessed raw demonstrations. \cite{li2025bilevel} proposes a bi-level learning architecture to mirror the structure of the bi-level planning, which focuses on the invention of neural relational predicates, assuming actions are predefined. Neuro-symbolic skill discovery, which involves abstracting continuous low-level state-action spaces \cite{keller2025neuro,shao2024learning} into discrete symbols to decompose a task into more manageable sub-tasks, has importance beyond its application in bi-level planning. \cite{keller2025neuro} uses neuro-symbolic policies to abstract states into predicates and a set of operators that define a transition function, which together define a planning problem that can be solved using PDDL. 
\cite{chitnis2022} proposed neuro-symbolic relational transition models where a neural network is responsible for low-level operator search in a given parametric space. 
\cite{wang2018,wang2021} proposed a sampling method for creating a set of potential action parameters along with the skills. Given a goal and learned parametric motion primitives, their system generates a plan using their PDDLStream framework \cite{garrett2018}.
\cite{shao2024learning} creates pseudo-labels derived from an expert-provided knowledge base formulated as a finite state machine. \cite{shah2024reals} proposed an approach to automatically learn relational and generalizable representations of actions and states from unlabeled demonstrations. On the other hand, our framework proposes a single generic end-to-end conditional neural architecture capable of discovering high-level skills from a few unlabeled low-level trajectories and generating low-level trajectories of these high-level skills given goals to generate and execute the low- and high-level plans.

Large language models are widely used for robotic planning due to their extensive world knowledge and ability to reason through tasks \cite{Huang2022zeroshotplan, huang2023inner}, making them ideal for complex scenarios. \cite{wei2022chain} demonstrates that even simple chain-of-thought prompting enables LLMs to perform complex reasoning tasks. Research in this field investigates using LLMs with prompting techniques to generate intermediate representations like predicate-operator pairs \cite{yang2024guiding} or code \cite{han2024interpret} from language instructions, which then directs low-level robot control. \cite{liang2023code, huang2023voxposer} uses LLMs to generate Python codes as policies via few-shot prompting. \cite{silver2024generalized} focuses on generating plans in the PDDL domain using chain-of-thought summarization. However, more recent studies like \cite{black2024pi_0,kim2024openvla} explore their usage as direct action policies (Vision Action Model) without using any intermediate steps. \cite{zawalski2024robotic} claims that the direct mapping between observations and low-level actions makes VLAs harder to generalize to out-of-dataset tasks and suggests implementing multiple reasoning steps before action prediction. Our paper uses VLMs to label demonstrations and generate high-level plans out of these labeled demonstrations. Before generating a high-level plan, we prompt it with multiple questions about the environment and object locations to maximize its reasoning potential.



\section{Method}
\subsection{Model}
\begin{figure*}[b]
    \centering
    \includegraphics[width=0.9\linewidth]{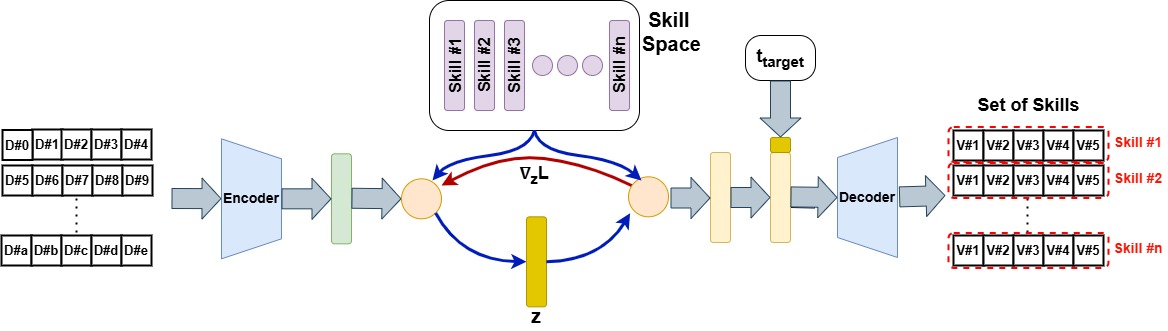}
    \caption{The Overview of our Model}
    \label{fig:single-channel-model}
\end{figure*}

\label{section:method}
This study proposes a novel neural network model that can learn discrete latent representations from skill demonstrations. The model utilizes the vector quantization approach \cite{van2017neural} to learn single discrete vectors from different variations of a high-level skill. For example, taking a sandwich from the refrigerator is a high-level skill, while the location in the refrigerator from which the sandwich is taken constitutes the variation or the low-level component of the high-level skill. Given a demonstration dataset that includes different variations of multiple high-level skills using a vector-quantized autoencoder architecture, different skills get assigned to different vectors in the vector space as training progresses. While learning the skill vector embeddings, the model also learns a distribution for each vector embedding that spans each high-level skill's low-level planning space. Furthermore, our model can also group unlabeled demonstrations into skills since the given demonstrations are given without labels and order. \hakann{In this context, the term label means which demonstration belongs to which skill.}

\begin{figure*}[t]
    \centering
    \includegraphics[width=0.7\linewidth]{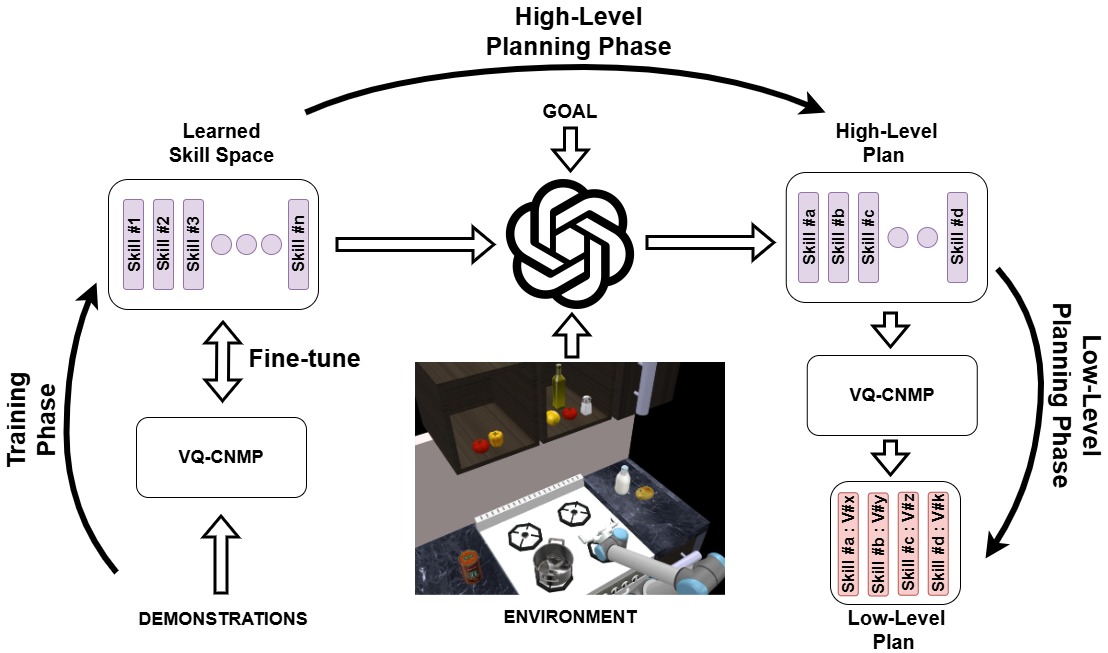}
    \caption{The detailed training, plan generation, and plan execution pipeline. }
    \label{fig:planning-system}
\end{figure*}

We used the architecture of the Conditional Neural Movement Primitives \cite{seker2019conditional} to leverage its ability to learn latent spaces from continuous movement trajectories. We also incorporated vector quantization to learn a discrete vector space from given demonstrations.\hakann{The model overview can be seen in Figure \ref{fig:single-channel-model}. The model takes sensorimotor trajectories (in this case, end-effector pose) as input and learns to reconstruct the same trajectory (similar to an Autoencoder) given random sample points from the trajectory. During training, the vector quantization bottleneck learns discrete representations for demonstration trajectories that belong to the same high-level skill. At each iteration, several sampled points from a selected trajectory are fed through the encoder to construct a latent vector. After the vector quantization layer, the time step of the desired output point on the trajectory ( $t_{target}$ ) is concatenated with the vector and passed through the decoder to get the output. The model is expected to output the part of the trajectory of the concatenated time step.} More formally, during training, given a set of demonstrations $D = {(t,SM(t))}$ where $SM(t)$ denotes sensorimotor information at time $t$, at each training iteration, a demonstration trajectory, $D_j$, is randomly sampled from the dataset. From this trajectory, a set of points is sampled ,which can be denoted as $(t_i,SM(t_i))_{i=0}^{i=n}$ and are fed through the encoder,
\begin{equation} \label{eq:1}
z_{i} = E((t_i,SM(t_i))|\theta) \quad (t_i,SM(t_i)) \in D_j
\end{equation}
where $z_i$ denotes the latent representation generated by the encoder $E$ with parameters $\theta$ using data point $(t_i,SM(t_i))$. Then these representations are averaged,
\begin{equation} \label{eq:2}
z_e = \frac{1}{n} \sum_{i=0}^{n} z_{i} 
\end{equation}
to get a single latent representation $z_e$. Then, similar to \cite{van2017neural}, the representation is passed through a discretization bottleneck by mapping it to the nearest vector in the skill space,
\begin{equation} \label{eq:3}
z_q = v_k \quad where \quad k = argmin_m(||z_e - v_m||) 
\end{equation}
where $v_m$ denotes the $m^{th}$ vector in the space. Then, the resulting representation $z_q$ is concatenated with the target time $t_{target}$ (the time step of the outputted data point) and  passed through the decoder to generate the output,
\begin{equation} \label{eq:4}
(\mu_{t_{target}},\sigma_{t_{target}} ) = Q((z_q,t_{target})|\phi) 
\end{equation}
where $\phi$ denotes the parameters of the decoder $Q$, $\mu_{t_{target}}$ denotes the mean and the $\sigma_{t_{target}}$ denotes the variance of the output. The loss of the model is defined as the following,
\begin{equation} \label{eq:5}
Loss =   -\log P(SM(t_{target})|\mu_{t_{target}},\sigma_{t_{target}}) + ||sg(z_e)  - z_q||^2_2 + \beta * ||z_e - sg(z_q)||^2_2 
\end{equation}
which includes both the reconstruction loss (negative-log likelihood) of the CNMP (the term on the left) and the loss terms of Vector Quantization (on the middle and the right). $sg$ denotes the stop gradient operator. Unlike \cite{van2017neural}, we observed that using $\beta$ values that are not small causes the gradients to explode. We used $\beta=0.25$ in our experiments and observed that $\beta$ values as small as 1 were enough to cause the gradients to explode. 
After the training, the learned skill embeddings include one vector for each high-level skill in the demonstration dataset, which can be used to generate variants of the learned high-level skills. Trajectories generated using these vectors are general representations 
In other words, each skill can be considered a trajectory distribution of its low-level trajectory variations encoded by the discovered mean vector. Using these vectors, a high-level skill repertoire (Learned Skill Space in Figure \ref{fig:planning-system}) can be formed, and an upper layer system can use the repertoire to make high-level plans. However, since we assume that the demonstration dataset is unlabeled, the skills the model discovers are also unlabeled. In other words, the system learns that a demonstration, $m$ is a variation of skill, $n$ for a set of given (but not labeled) demonstrations and skills. As such, what each skill $n$ corresponds to for a higher-level planner is unknown. Labeling these vectors (what each skill is, like opening the drawer) can be done by a human partner (i.e., by watching the execution), or a Multi-Modal LLMs can be used to automate the processing of our system.

\subsection{Self-Supervised Learning} \label{subsec:self-sup}

While the model can successfully cluster the demonstrations into discrete skills, the actions that the system generates are not always reliable for use in low-level plans. To remedy this issue,
we propose self-supervised learning using the skill labels the system discovers. After the initial training is complete and the skills are discovered, i.e., the demonstrations are clustered into skills, we train a new model from scratch using the discovered skill labels, significantly increasing the low-level planning performance. To achieve this, instead of letting the system decide which vector to choose by Euler distance in the vector quantization layer, we used the discovered cluster labels to assign the demonstrations to the same vector in the first training phase.
\subsection{Planning} \label{subsec:planning}
After the skill space is learned and labeled, for making high-level plans,  the actions are given in a prompt along with the goal and the image of the environment (Figure \ref{fig:planning-system}). The prompt instructs that only the given actions can be used and an ordered list should be returned. 
For the low-level plan, an input high-level skill vector is modified using gradient descent as in \cite{aktas2024multi} to generate the desired goal prediction at the decoder output.
For practicality, we used end-effector pose data and gripper state as input to the model for the planning experiments, so that the part of the action prediction where the robot makes contact with the object would be close to the object's location, which is what differs between the low-level versions of the same high-level skill. For each step of the high-level plan, the vector of that step's skill is given as input to the decoder of our model, along with the target time close to the time the robot makes contact with the object it is interacting with. Since the change between the variations of the skills is where they retrieve the objects from, the time step that the robot makes contact with the object will be close to the initial location of the object for that generated trajectory. After that, the difference between the predicted object location and the actual object location is calculated using MSE.
\begin{equation}
    loss = MSE((x_{\mu_{e_{t}}},y_{\mu_{e_{t}}},z_{\mu_{e_{t}}}),(x_{object},y_{object},z_{object}))
    \label{eq:plan3}
\end{equation}
The result is used as loss, and the gradient of the input vector is taken and applied. Note that the model's parameters are unchanged in this step, and only the vector given as input to the decoder is adjusted. This process is repeated until the predicted object location is sufficiently close (empirically set, based on the object, for all less than 2 cm). After the process is complete, the adjusted vector is fed into the system's decoder along with all time steps to generate the action, which is the low-level variation of this skill to achieve this step of the high-level plan. This process is repeated for each vector in the high-level plan.



\section{Experimental Results}\label{section:experiments}
We evaluated the performance of our approach both in simulated and real-world experiments. For simulated experiments, we created a kitchen environment that includes everyday kitchen items such as salt and oil, as well as common ingredients like tomatoes, peppers, lemons, and potatoes.  We divided the environment into four regions of interest (left and right cupboards, left and right of the stove). We collected demonstrations that achieved pick-up and place tasks from different initial object locations for pick-up and different final positions for place. Note that the collected demonstrations only span a small part of the overall manipulation areas. For all actions, the robot starts from the same initial position and completes the action in the same position, allowing for continuous execution. For example, for pick-up actions, the robot starts from its initial position with an open gripper, reaches the specified object in the area of interest, picks it up, and returns to its initial position. We formulated the actions with predicates for high-level planning as $ ActionType(object,areaOfInterest)$.
For instance, picking up a tomato from the right cupboard is formulated as $pickup(tomato,rightCupboard)$. However, it should be noted that the skills are not object-specific. The same action used to pick up a tomato from the same position can be used to pick up the oil bottle. In addition to pick-up and place skills defined in the four areas of interest, we also included pouring and placing into the pot actions. However, while we included the demonstrations of these skills in the training set, since only four demonstrations span the entire execution space of these skills (the skills assume that the object to be poured/placed is already grasped, and there are only four places the pot can be) we did not include these skills to the low-level evaluation. We utilized DINO \cite{caron2021emerging} to get the locations of the objects in the environment for our low-level planner to use. To demonstrate that our model can also be deployed in real-world settings, we conducted real-world experiments, including the following tasks: loading glasses and plates into a dishwasher from varying object locations, making coffee using a coffee machine, and watering plants. For all simulated and all real-world experiments, we used a UR-10 arm with a BarrettHand gripper and XArm 7, respectively.   
\subsection{Skill Discovery}

\begin{table}[t]
    \centering
    \begin{tabular}{c|ccccc}
 & \multicolumn{5}{c}{Skill Space Size}\\  
         &  3&  5&  10&  20&  100\\ \hline 
         perfect clustering&  73/100&  27/100&  30/100&  32/100&  16/100\\ 
         Accuracy&  0.98292&  0.97
&  0.95876&  0.94720&  0.89894

    \end{tabular}
    \caption{Demonstration Clustering results. Perfect clustering represents the number of models where demonstrations of each skill were assigned to the same vector. Accuracy measures the percentage of demonstrations of the same skill assigned to the same vector. }
    \label{tab:clustering}
\end{table}

To test the skill discovery performance of our model, we trained 100 randomly initialized models and analyzed them in folds of 10.  The results showed that in \hakan{27} of all trials, the skills were clustered perfectly into the five vectors that were in the high-level skill space (meaning all demonstrations of the same skill have been mapped to the same skill vector embedding by the model). We further analyzed the results in folds of 10 to find a practical way of selecting a model that automatically clustered the demonstrations. In all batches, we observed that two of the three models with the least valued losses in the batch were perfectly clustered. 

 
We also investigated how the model behaves when the size of the skill space is more or less than the number of skills in the demonstrations. We set the size of the skill space (the number of embeddings in the vector quantization layer) to 3, 10, 20, and 100, trained 100 models each, and evaluated them the same way. Results showed that when the size of the vector space is 3, the model clustered some of the skills together, and out of the 100 models, 73 of them were perfectly clustered. We also observed that when the skill space is smaller, the combined loss values of the models were significantly higher. When the skill spaces were larger than the skills in the demonstration set, we observed that 30, 32, and 16 of the 100 models were clustered perfectly for the models with the vector spaces of size 10, 20, and 100, respectively. When we examined the demonstration distribution of each model, we observed that the model divided the demonstrations that belong to the same skill into different vectors, reducing the vector quantization loss. In practice, to find a good estimation of the number of skills in the demonstration dataset, one can train several folds of models with varying vector space sizes and compare the vector quantization losses. We also calculated an accuracy value by setting the correct label of each skill to the vector to which the highest number of demonstrations are mapped. Hence, in this context, accuracy shows how clustered demonstrations of the same skill are, and whether it is high or low is not an indication of how well the model behaves. As seen in Table \ref{tab:clustering}, as the number of vectors in the vector space increases, the accuracy values drop, meaning the demonstrations belonging to the same skill get distributed to multiple vectors, which, as stated, reduces the vector quantization losses.


Based on the results, in practice, our model can be utilized by training a fold of models with significantly larger skill spaces than the number of skills in the demonstration dataset, and the ones with the least losses can be used \hakann{for skill discovery}. The vectors that represent the same action can be detected and labeled by executing the actions generated by the decoders using the vectors. \hakann{After the discovery step is complete, a new model with vector space size equal to the number of discovered skills can be trained using the self-supervised training scheme described in subsection \ref{subsec:self-sup}.}
\begin{table}[t]
    \centering
    \scalebox{0.8}{
    \begin{tabular}{c|llll}
         Task Type & GPT w/o DINO & GPT+DINO\textsubscript{GPT} & GPT+DINO\textsubscript{Gemini} & Gemini+DINO\textsubscript{Gemini} \\
         \hline
         Short ($\leq$2 actions)   & 93.0  & 92.67 & 96.0  & \textbf{97.19} \\
         Medium (3–7 actions)      & 92.29 & 91.71 & 97.71 & \textbf{97.96} \\
         Long ($ > $7 actions)     & 48.28 & 47.14 & 56.0  & \textbf{65.18} \\
         \hline
         All tasks       & 77.10 & 76.40 & 82.60 & \textbf{86.10} \\
    \end{tabular}}
    \caption{Task performance grouped by action sequence length. Tasks are categorized as short ($\leq$2 actions), medium (3–7 actions), and long ($> $7 actions).}
    \label{tab:planning_performance}
\end{table}
\begin{table}[b]
    \centering
    \scalebox{0.85}{
    \begin{tabular}{clccccc|clll}
 & & & \multicolumn{4}{c}{unsupervised}& \multicolumn{4}{c}{self-supervised}\\
          & &&  1&  2& 4& 10&1&2& 4&10\\\hline
 \multirow{4}{*}{simulated}&  \multirow{2}{*}{single-task}&pick-up& \%0& \%0& \%5& \%20&\%60& \%100& \%100&\%100\\
         & &place&  \%0&  \%0&  \%10& \%15&\%50& \%90& \%100&\%100\\
 &  \multirow{2}{*}{multi-task}&retrieve& \%0& \%0& \%5& \%5& \%45& \%85& \%100&\%100\\
 & & cook& \%0& \%0& \%0& \%0& \%30& \%80& \%90&\%85\\\hline
         \multirow{4}{*}{real}& \multirow{3}{*}{single-task}&dishwasher-glass&  \%0&  \%0&  \%0& \%5&\%50& \%90& \%100&\%90\\
  & &dishwasher-plate& \%0& \%0& \%5& \%0&\%65& \%95& \%100&\%100\\
 & & water-plant& \%0& \%0& \%10& \%10& \%70& \%100& \%100&\%100\\
 & multi-task& make-coffee& \%0& \%0& \%0& \%0& \%20& \%75& \%85&\%80\\
    \end{tabular}}
    \caption{Comparison of our model's low-level planning performance when trained self-supervised vs unsupervised using varying number of demonstrations per skill.}
    \label{tab:low-level}
\end{table}
\subsection{Labeling using Multi-Modal LLMs}\label{subsec:labeling}

\begin{wraptable}{r}{0.55\textwidth}
    \centering
    \begin{tabular}{c|cl|ll}
 & \multicolumn{2}{c}{$4\times2$}& \multicolumn{2}{c}{$4\times4$}\\
 
         Action&   Gemini& GPT& Gemini&GPT\\
         \hline
         Pick& 
     \%34.38& \%59.37& \%31.25&\textbf{\%65.62}\\
 Place& \%6.25& \%3.12& \%9.38&\textbf{\%15.62}\\\end{tabular}
    \caption{LLM labeling accuracy results by action, model, and grid size.}
    \label{tab:LLM}
\end{wraptable}

To test whether Multi-Modal LLMs can be used for labeling skills, we took image snapshots of the execution of the action generated using the vectors in the learned skill space at different time steps. To this end, we first took a snapshot from the middle of the action execution, gave it to the LLMs, and asked them to analyze the environment. We then used the analysis alongside a grid of execution snapshots taken from different parts of the execution to the LLMs and asked for the action type. We experimented with varying viewing angles for each task to make the object of interest as visible as possible and reported the best mean results. We experimented with Gemini 2.5 Flash and GPT-4.1. The validation of the predictions is also done using GPT-4.1 by providing the model prediction and ground truth. The validator either accepts or rejects the predictions and outputs a confidence value, which we also used while manually inspecting the results. Results for pick and place actions can be seen in Table \ref{tab:LLM}. One interesting finding is that although we annotated the images to show the execution order of the snapshots, the models tended to output pick even for the place actions, which is why the results for place are significantly worse than the ones for the pick action. We also experimented with different numbers of images in the grid, which showed that while the performance of the GPT model increased with more images, the performance of Gemini was inconsistent. Manual inspection showed that a further decrease in the number of images renders the sequence uninterpretable, while increasing it is not viable due to a trade-off between grid image size and token count.
\vspace{-5mm}

\subsection{Planning Performance}

\subsubsection{High-Level Planning Performance}
\begin{wrapfigure}{l}{0.4\textwidth}
    \centering\includegraphics[width=1\linewidth]{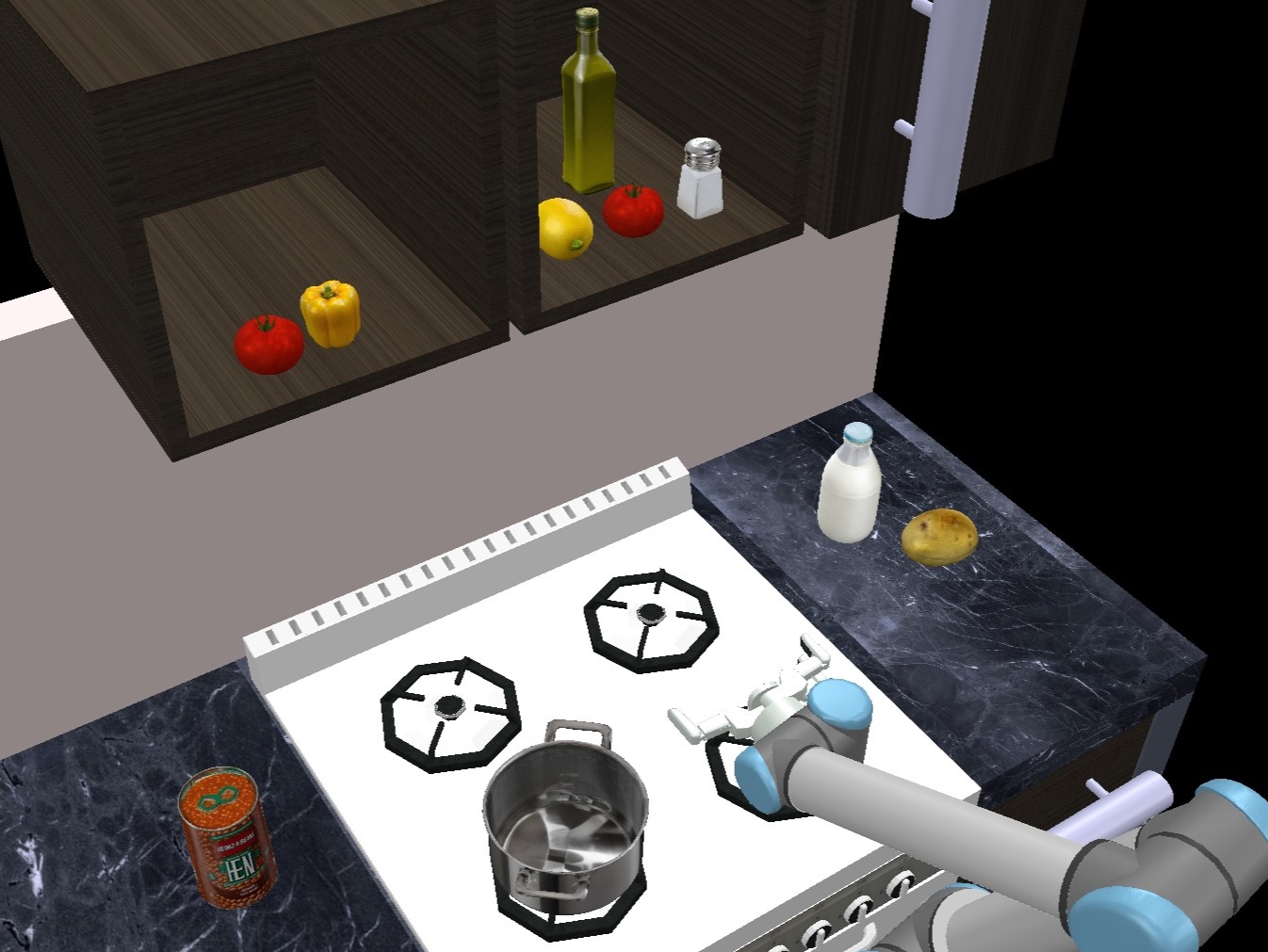}
\caption{Image of the environment used for high-level planning.}
\label{fig:environment}
\end{wrapfigure}

To utilize the LLMs for high-level planning, we first provide the initial state of the environment to the LLM and make it analyze the scene and list the objects in the scene. Using this list, we used DINO to annotate and locate each image (which we later used in low-level planning). Finally, we give the environment analysis alongside the environment image to the LLMs and ask it to make a plan using the provided high-level skills that would achieve the given task. Across all trials, GPT-4.1 mini and Gemini are used for the final planning step. We observed no significant performance difference between GPT-4.1 and GPT-4.1 mini at this step; thus, we chose the latter for efficiency. The resulting plan is given to an LLM for validation. Validator takes the goal specification (manually crafted success criteria for the goal) and is prompted to check whether the provided plan satisfies the goal. Similar to the previous section, the validator outputs the decision and confidence level used for evaluation. For the evaluations, 10 different environment settings viewed from 5 different angles are used. We created 20 tasks of varying degrees of complexity (details can be found in the Appendix) and evaluated all models on the same tasks. The summarized results can be seen in Table \ref{tab:planning_performance}. We experimented with annotating environment images using DINO and found that annotation quality significantly impacts performance. Gemini produced more accurate and comprehensive object lists, improving task performance, whereas GPT exhibited frequent misclassifications, which led to degradation. We will refer to these experiments as DINO\textsubscript{Gemini} and DINO\textsubscript{GPT}.
\vspace{-3mm}

\subsubsection{Low-Level Planning Performance}
\hakann{In this section, we trained models with a number of vector embeddings equal to the number of skills using the self-supervision technique described in Subsection \ref{subsec:self-sup}.} We evaluated both single and multi task performance by setting goals that can be achieved by executing only one skill and more for single and multi task respectively. For the multi-task trials we conducted, retrieval was achieved by retrieving an object behind another object (hence the object on the front needs to be removed first), and cooking tasks, which are defined by adding various ingredients to the pot. In the real-world experiment, we set a make-coffee goal where the model is expected to pick up a cup, put it in the coffee machine, press the button to make the coffee, and bring the cup back to another position. Dishwasher tasks are defined as picking up glasses or plates from varying locations and loading them into a dishwasher and watering plants is defined as picking up a water bottle from different locations and pouring it into a flowerpot. We experimented with both the unsupervised and the self-supervised cases. We conducted 20 trials per experiment where we changed object positions to evaluate generalization. The results shown in the Table \ref{tab:low-level} indicate that the model that was trained using self-supervision significantly outperformed the unsupervised one, and the self-supervised model was able to generalize well in all experiments when provided with two or more demonstrations per skill.

\vspace{-3mm}
\section{Conclusion}
\label{section:conclusion}
\vspace{-2mm}
In this paper, we propose a novel neural network model capable of clustering skills in mixed datasets and that can be used for bi-level planning. We tested its ability to discover skills under different conditions, such as varying sizes of the skill space. We also tested whether LLMs can be used to label the discovered skills. Lastly, we show our system's planning performance. As a future work, a promising research direction would be to incorporate effect predicates into the model and the pipeline, which would allow the model to do error detection and may result in better performance in long-horizon plans.



\section{Limitations}\label{sec:limitations}

One limitation of our model is that demonstrations must follow the same distribution to be classified as having the same skill. Furthermore, since the system depends on an external agent for reasoning and we \erhan{do not} use state abstractions, its planning performance highly depends on the external agent's perception and high-level planning capabilities. Additionally, while the environment image 
\erhan{contains sufficient} information about the initial state for high-level planning, the exact locations of the objects to be interacted with are needed to make the low-level plans.  Finally, like many imitation learning models, the setup assumes expert-level demonstrations are provided to the system.


\clearpage


\bibliography{references}  

\appendix

\begin{table}
    \centering
\scalebox{0.8}{
\begin{tabular}{c|c|cccc}
Ingredient & Action & ($4\times4$) Gemini & ($4\times4$) OpenAI & ($4\times2$) Gemini & ($4\times2$) OpenAI \\ \hline
beans & Pick & 50.0 & 100.0 & 25.0 & 50.0 \\
beans & Place & 25.0 & 25.0 & 0.0 & 0.0 \\
pepper & Pick & 50.0 & 75.0 & 25.0 & 75.0 \\
pepper & Place & 0.0 & 25.0 & 0.0 & 0.0 \\
lemon & Pick & 0.0 & 50.0 & 50.0 & 75.0 \\
lemon & Place & 25.0 & 0.0 & 0.0 & 0.0 \\
milk & Pick & 0.0 & 25.0 & 25.0 & 0.0 \\
milk & Place & 0.0 & 0.0 & 25.0 & 0.0 \\
oil & Pick & 75.0 & 75.0 & 75.0 & 75.0 \\
oil & Place & 25.0 & 25.0 & 0.0 & 0.0 \\
potato & Pick & 25.0 & 100.0 & 50.0 & 100.0 \\
potato & Place & 0.0 & 25.0 & 25.0 & 0.0 \\
salt & Pick & 25.0 & 0.0 & 0.0 & 0.0 \\
salt & Place & 0.0 & 0.0 & 0.0 & 0.0 \\
tomato & Pick & 25.0 & 100.0 & 25.0 & 100.0 \\
tomato & Place & 0.0 & 25.0 & 0.0 & 25.0 \\
\end{tabular}}
\caption{LLM labeling accuracy results by ingredient, action, model, and grid size.}
\label{tab:LLM_app}
\end{table}

\section{Experimental Results}

\subsection{Labeling Using Multi-Modal LLMs}

Results for all objects can be seen in Table \ref{tab:LLM_app}. The prompt we used to detect environment can be seen below.

\lstset{basicstyle=\ttfamily,
frame=single,
breaklines=true,
}

\begin{lstlisting}
You are an expert environment detection assistant.  
The user will provide you with a frame captured from a robot's perspective.  
Your task is to:
1. Carefully analyze the frame to understand the surroundings and identify all visible objects.
2. For each object, guess its exact type and describe its appearance, location, and relative position within the frame. If the image is already annotated, you can use the annotations to help you.
3. Synthesize your observations to deduce the overall environment the robot is operating in.

Format your response strictly as follows:

<analysis>
Detailed analysis of the frame here, listing objects, their locations, and descriptions.
</analysis>

<environment>
A concise, clear identification of the environment based on the analysis.
</environment>

Be thorough, logical, and base your conclusions only on the visual information provided.
\end{lstlisting}

The prompt we used to generate labels for the snapshots can be seen below.

\begin{lstlisting}
You are an assistant specialized in detecting a robot's skill based purely on visual input.  
The user will provide you with a series of frames showing a robot performing a basic skill.  
Your objectives are:
1. Analyze the frames carefully to understand the environment, identify objects, and observe the robot's precise actions and movements Also, use previous informations you have from your answers.
2. Describe only what is explicitly visible - such as the robot moving, reaching, grasping, or holding - without inferring intent or goals beyond what is directly observed.
3. Determine the robot's skill based strictly on the sequence of visible actions. If the robot has not completed an action (e.g., grasping an object), state exactly what is seen (e.g., "moving toward the object" rather than "grasping" or "placing").
4. Be specific about the objects involved.

Format your response exactly as:

<analysis>
Detailed description of the frames, environment, objects, robot's movements, and their sequence. Focus only on observed facts without interpreting future intentions.
</analysis>

<skill>
Specific description of the skill based solely on visible actions. Be factual and avoid assuming unobserved outcomes. Be specific about objects and actions.
</skill>

Only base your conclusions on what is fully visible in the provided frames. Do not infer missing steps, future actions, or unseen goals.
\end{lstlisting}

The prompt we used to validate the label can be seen below.
\begin{lstlisting}
You are an expert evaluation assistant.  
Your task is to compare a model's predicted skill description to the ground truth skill description.  
Follow these steps:
1. Carefully read the ground truth and the model's output.
2. Determine whether the model's output is 'correct' (accept) or 'incorrect' (reject) based on semantic meaning, not exact wording.
3. Provide a short explanation of your reasoning.
4. Assign a confidence level
    - High: You are very confident in your evaluation (whether accepting or rejecting). Exact or very clear mismatch/match.
    - Medium: Mostly correct but with some uncertainty or small differences.
    - Low: Significant uncertainty or ambiguity about whether it matches.

Format your response exactly as:

<explanation>
Your short explanation here.
</explanation>

<decision>
Accept or Reject
</decision>

<confidence>
High / Medium / Low
</confidence>


Important:
- Focus purely on semantic meaning.
- Be concise and accurate.
- Always fill all three sections: explanation, decision, confidence.

## Examples:

Ground truth: "pick up the bottle"  
Model output: "grab the bottle"

<explanation>
The model's output "grab the bottle" accurately captures the same meaning as "pick up the bottle."
</explanation>

<decision>
Accept
</decision>

<confidence>
High
</confidence>


Ground truth: "place the cup on the left counter"  
Model output: "move the cup to the table"

<explanation>
The model describes moving the cup to the table instead of placing it on the left counter. While the action type (moving/placing) is correct, the locations are different. However, since a table and a counter could be visually or conceptually similar, the confusion is understandable, leading to rejection with medium confidence.
</explanation>

<decision>
Reject
</decision>

<confidence>
Medium
</confidence>


Ground truth: "pick up the bowl"  
Model output: "put the bowl on the stove"

<explanation>
The model output describes placing the bowl somewhere, while the ground truth requires picking it up first, indicating a major difference in action type.
</explanation>

<decision>
Reject
</decision>

<confidence>
High
</confidence>

\end{lstlisting}

\subsection{Planning}
\subsubsection{High-Level Planning Performance}

\begin{table}[t]
    \centering
\scalebox{0.7}{
\begin{tabular}{l|cccc}
\makecell[l]{Task Name} & GPT w/o DINO & GPT+DINO\textsubscript{GPT} & GPT+DINO\textsubscript{Gemini} & Gemini+DINO\textsubscript{Gemini} \\ \hline
Pick up an oil bottle & \textbf{\%100.00} & \%98.00 & \textbf{\%100.00} & \textbf{\%100.00} \\
Pick up a tomato & \%98.00 & \%98.00 & \textbf{\%100.00} & \textbf{\%100.00} \\
Pick up a potato & \%96.00 & \textbf{\%98.00} & \textbf{\%98.00} & \%95.92 \\
Pick up a lemon & \%74.00 & \%76.00 & \%92.00 & \textbf{\%100.00} \\
Pick up salt & \textbf{\%100.00} & \textbf{\%100.00} & \textbf{\%100.00} & \textbf{\%100.00} \\
Pick up a bean can & \textbf{\%90.00} & \%86.00 & \%86.00 & \%87.23 \\
Pour oil into the pot & \textbf{\%100.00} & \textbf{\%100.00} & \textbf{\%100.00} & \textbf{\%100.00} \\
Pour salt into the pot & \textbf{\%100.00} & \textbf{\%100.00} & \textbf{\%100.00} & \textbf{\%100.00} \\
Place a tomato into the pot & \textbf{\%100.00} & \textbf{\%100.00} & \textbf{\%100.00} & \textbf{\%100.00} \\
Place a potato into the pot & \textbf{\%98.00} & \textbf{\%98.00} & \textbf{\%98.00} & \%96.00 \\
Place a lemon into the pot & \%76.00 & \%76.00 & \%94.00 & \textbf{\%100.00} \\
Place a lemon in the \\ left cupboard & \%72.00 & \%68.00 & \%92.00 & \textbf{\%93.88} \\
Place an oil bottle to the \\ left of the stove & \textbf{\%100.00} & \textbf{\%100.00} & \textbf{\%100.00} & \%95.83 \\
Place all pourable objects \\ in the right cupboard & \%22.00 & \%22.00 & \%12.00 & \textbf{\%34.69} \\
Empty the right cupboard & \%36.00 & \%24.00 & \%32.00 & \textbf{\%36.73} \\
Place all sour objects \\ to the left of the stove & \%54.00 & \%60.00 & \%82.00 & \textbf{\%95.45} \\
Make stew including: \\ tomato, lemon & \%66.00 & \%68.00 & \%76.00 & \textbf{\%87.23} \\
Make stew including: \\ tomato, lemon, and potato & \%58.00 & \%70.00 & \%68.00 & \textbf{\%72.00} \\
Make stew including: \\ tomato, lemon, salt, oil & \%28.00 & \%18.00 & \%38.00 & \textbf{\%47.92} \\
\makecell[l]{Make stew including: tomato, lemon, salt, and oil. \\ Empty the left side of the counter first, place all \\ ingredients there, and then make the stew.} & \%74.00 & \%68.00 & \textbf{\%84.00} & \%82.22 \\
\hline
\textbf{Overall Accuracy} & \%77.10 & \%76.40 & \%82.60 & \textbf{\%86.10} \\
\end{tabular}}
\caption{Planning and task performance across different configurations and models.}
\label{tab:planning_performance_app}
\end{table}

The table including results of individual tasks in this subsection can be seen in Table \ref{tab:planning_performance_app}. The prompt template we used to generate list of objects can be seen below.

\begin{lstlisting}
You are a helpful assistant that identifies and lists all visible objects from an image.  
Your task is to:
1. List all detected objects using detailed, specific names (e.g., "a red mug" instead of just "a mug").
2. Start each object with an indefinite article ("a" or "an") if it appears once, if multiple instances are visible start with "many" instead of indefinite article and include the object only once.
3. List each object separately, even if they are similar or related. Do not group objects together (e.g., write "a spoon, a fork" instead of "a spoon and a fork").
4. Separate the objects with commas.
5. Do not include any extra text, explanations, or formatting-only output the comma-separated list.
\end{lstlisting}

The prompt template we used to make the high-level plans can be seen below.
\begin{lstlisting}
You are an assistant specialized in creating detailed action plans for a robot using a predefined action set.  
The user will provide you with the robot's initial state and the desired goal with an annotated image. The user will also provide a list of detected objects in the environment. These objects may or may not be fully accurate, but you should use them to help guide your analysis and planning while still relying primarily on visible evidence.

Your task is to:
1. Analyze the initial state and goal based only on the visible information provided. You cannot use any external knowledge.
2. Plan a step-by-step sequence using only the following allowed actions:
   - Pick: The robot picks up an object from a location and returns to its original position while holding the object.
   - Place: The robot places a held object at a target location and returns to its original position without the object.
   - Pour: The robot pours from a held object into a target location and returns to its original position still holding the object. Note: The robot must pick the object before pouring, and it keeps holding the object after pouring.
3. Accessible locations are: "left cabinet", "right cabinet", "left of the stove", "right of the stove", "pot".
4. Each step must clearly mention the action (Pick, Place, Pour), the object involved, and the target location if applicable, based strictly on what is visible.
5. You must only use the allowed actions (Pick, Place, Pour) and must follow their defined behaviors precisely. No other actions are allowed.
6. After completing all action steps, end the plan by writing Done as the final numbered step.
7. Be aware some objects may be in the way of others, and the robot must move them to access the target object.

Format your response exactly as:

<analysis>
Your detailed analysis of the initial and goal states here.
</analysis>

<plan>
1. Action: Description
2. Action: Description
...
n. Action: Description
n+1. Done
</plan>

Be logical, detailed, and strictly base your plan only on observable facts.
\end{lstlisting}

The prompt we used to validate the generated plans can be seen below.

\begin{lstlisting}
You are an expert evaluation assistant.  
Your task is to evaluate whether a given plan correctly achieves a specified goal, based on a provided goal specification.

The user will provide:
- A goal (natural language description of the intended outcome).
- A goal specification (a breakdown of the essential steps or effects that the plan must satisfy).
- A plan (a list of actions).

Your evaluation steps:
1. Read and understand the goal and goal specification.
2. Check whether the plan satisfies all the conditions in the goal specification, regardless of order.
3. Accept the plan if:
   - It includes all the required steps from the goal specification.
   - It avoids actions that conflict with the goal (e.g., adding unneeded ingredients).
   - Extra, harmless steps (e.g., picking and placing an unused item) are present but do not interfere with the goal.
   - The actions are limited to `pick`, `place`, `pour`, and `Done`.
4. Reject the plan if:
   - Any required part of the goal specification is missing.
   - The plan includes **disruptive or contradictory actions**.
   - The plan violates the goal in a critical way.

Even if the plan is not a complete task solution, you should focus your judgment on whether it matches the goal specification.

Then output:
- A short explanation of your reasoning.
- A decision: `Accept` or `Reject`.
- A confidence level:
   - High: You are very confident in your evaluation.
   - Medium: There is some ambiguity but your judgment is mostly certain.
   - Low: The case is unclear or borderline.

Format your response exactly as:

<explanation>
Your short explanation here.
</explanation>

<decision>
Accept or Reject
</decision>

<confidence>
High / Medium / Low
</confidence>


Important:
- Confidence reflects how certain you are about your judgment, not how "complete" the plan is.
- Do not penalize extra steps unless they contradict the goal.
- Focus strictly on the match between goal specification and **plan behavior.
\end{lstlisting}

The prompt we used to map the steps in natural language to our formal language can be seen below.

\begin{lstlisting}
You are an assistant that maps a plan in natural language into a formatted sequence of actions.  
The user will provide you with a plan described in natural language.  
Your task is to:

1. Analyze the plan carefully and break it down into detailed steps, matching the intended robot actions.
2. Use only the following actions: `pick`, `place`, `pour`.
3. Format each step in the following style:  
`action(object,location)`

   - `action` must be lowercase.
   - `object` must come from the known list of objects.
   - `location` must be one of: "left cabinet", "right cabinet", "left of the stove", "right of the stove", "pot".

In the `<plan>` section:
- Only output the formatted action lines without adding any extra natural language.
- Each action must strictly match the style: `action(object,location)`.
- Location must exactly match one of the specified locations.
- Do not explain the actions inside the `<plan>`.
- After the last action, write `n. Done` where `n` is the next number.

Format your response exactly as:

<analysis>
Your detailed breakdown and reasoning here.
</analysis>

<plan>
1. action(object,location)
2. action(object,location)
...
n. Done
</plan>


Important:
- Follow the exact syntax strictly, matching the provided examples.
- Only base actions on the provided plan.
- Do not add assumptions, natural language sentences, or extra descriptions inside the `<plan>`.
- Only structured action calls are allowed inside the `<plan>`.
\end{lstlisting}

\end{document}